\documentclass[lettersize,journal]{IEEEtran}
\usepackage{amsmath,amsfonts}
\usepackage{algorithmic}
\usepackage{algorithm}
\usepackage{array}
\usepackage[caption=false,font=normalsize,labelfont=sf,textfont=sf]{subfig}
\usepackage{textcomp}
\usepackage{stfloats}
\usepackage{verbatim}
\usepackage{graphicx}
\usepackage[nocompress]{cite}
\usepackage{arydshln}
\usepackage{ragged2e}
\usepackage[
  colorlinks=true,
  linkcolor=blue,
  citecolor=blue,
  urlcolor=blue
]{hyperref}
\begin{document}

\newcommand{\appref}[1]{%
  \hyperref[#1]{Appendix~\ref*{#1}}%
}

\title{ReliCAD: From Uncertain LLM Generation to Reliable Parametric CAD Modeling}

\author{Peng Zheng, Xintong Dong, Chuanyang Li, Jiaxin Jing, Chuqi Han, Hailong Shen, Yanzhi Song, and Zhouwang Yang*
\thanks{P. Zheng, X. Dong, C. Li, J. Jing, C. Han, H. Shen,
Y. Song, and Z. Yang are with the University of Science and
Technology of China, Hefei, Anhui 230026, China
(e-mail:
pengzheng@mail.ustc.edu.cn;
weiss@mail.ustc.edu.cn;
lichuanyang@mail.ustc.edu.cn;
smallmatch@mail.ustc.edu.cn;
hanchuqi@mail.ustc.edu.cn;
shenhail@mail.ustc.edu.cn;
yanzhis@ustc.edu.cn;
yangzw@ustc.edu.cn).}
\thanks{Corresponding author: Zhouwang Yang
(e-mail: yangzw@ustc.edu.cn).}
}

% \author{Peng Zheng, Xintong Dong, Chuanyang Li, Jiaxin Jing, Chuqi Han, Hailong Shen, Yanzhi Song, and Zhouwang Yang*
% \thanks{P. Zheng, X. Dong, C. Li, J. Jing, C. Han, H. Shen,
% Y. Song, and Z. Yang are with the University of Science and
% Technology of China, Hefei, Anhui 230026, China
% (e-mail:
% pengzheng@mail.ustc.edu.cn;
% weiss@mail.ustc.edu.cn;
% lichuanyang@mail.ustc.edu.cn;
% smallmatch@mail.ustc.edu.cn;
% hanchuqi@mail.ustc.edu.cn;
% shenhail@mail.ustc.edu.cn;
% yanzhis@ustc.edu.cn;
% yangzw@ustc.edu.cn).}
% \thanks{Corresponding author: Zhouwang Yang
% (e-mail: yangzw@ustc.edu.cn).}
% }


\maketitle

\begin{abstract}
Large language models have shown considerable potential for natural-language-driven parametric CAD modeling. However, a fundamental contradiction exists between their probabilistic generation and the deterministic requirements of CAD modeling, resulting in limitations in reliability, design-intent preservation, and geometric validity. 
Existing methods typically rely on large-scale annotated datasets, lack explicit modeling of design intent, and underutilize the deterministic capabilities of CAD kernels. 
To address these limitations, we propose ReliCAD, a unified framework that transforms uncertain LLM generation into reliable parametric CAD modeling. Through explicit design-intent modeling, ReliCAD converts user requirements into structured design specifications and explicitly models geometric relations, topological dependencies, and feature construction order. It then generates constraint-aware parametric instructions and invokes the CAD kernel through an Agent-ready API to perform geometric construction and constraint solving. ReliCAD further records runtime evidence and employs a verification-feedback mechanism to assess consistency between the generated model and the design specifications, enabling error localization and iterative repair. Experiments on the public HistCAD generation dataset and our multi-granularity CAD editing dataset demonstrate that ReliCAD significantly outperforms baseline methods, achieving 99.8\% validity rate and 0.8753 IoU. ReliCAD provides a verifiable, repairable, and generalizable approach to natural-language-interactive CAD modeling.
\end{abstract}


\begin{figure}[htbp]
\centering
\includegraphics[width=\linewidth]{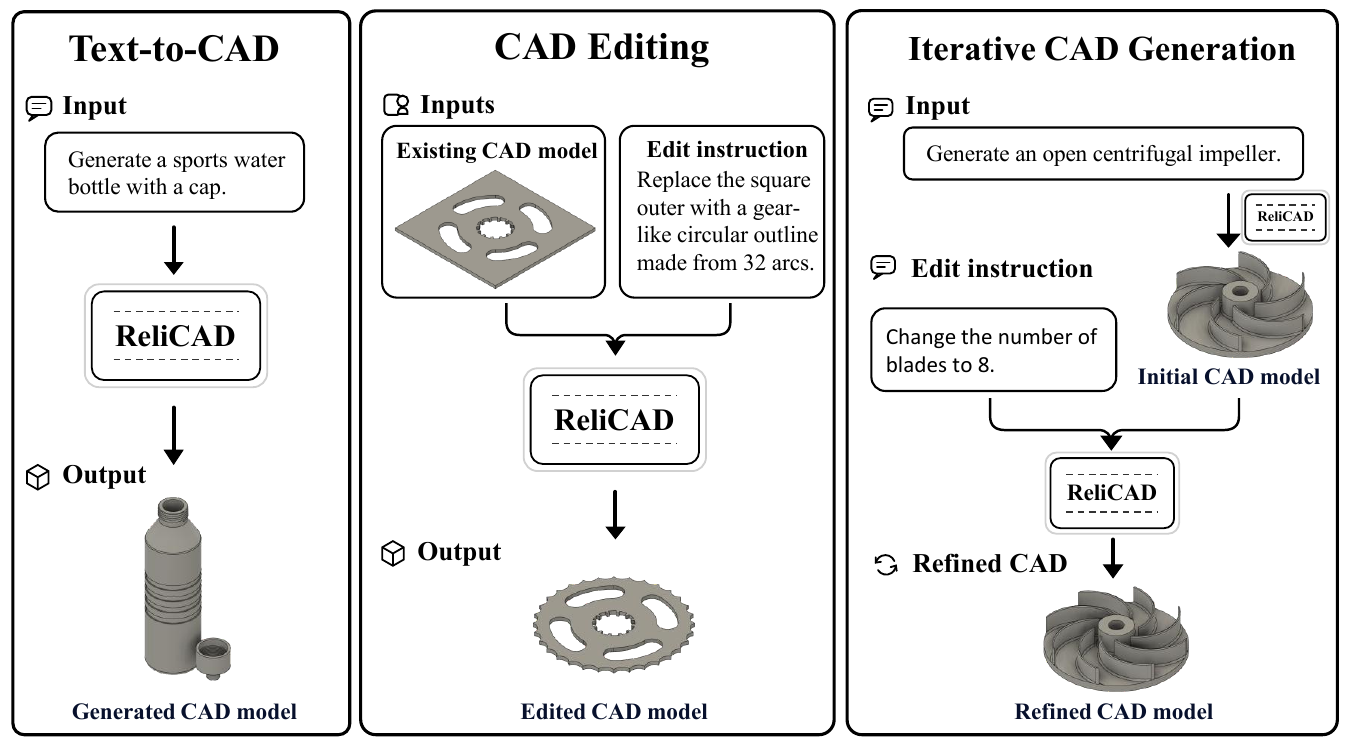}
\caption{Demonstration of CAD models generated and edited by ReliCAD under three interaction modes.}
\label{fig:three interaction modes}
\end{figure}

\section{Introduction}

\IEEEPARstart{P}{arametric} CAD modeling is a foundational technology in product design and engineering manufacturing. 
By combining constrained sketches, feature-based operations, and editable construction histories, it represents product geometry and preserves design intent for downstream modification and reuse \cite{camba2016parametric,seff2021vitruvion}. However, conventional CAD modeling still relies heavily on skilled engineers to manually construct precise models. This process requires substantial expertise and is often time-consuming, which limits its ability to support efficient modeling and rapid design iteration.

Recent advances in large language models have opened new possibilities for natural-language-driven CAD modeling \cite{brown2020language, achiam2023gpt, yao2022react, schick2023toolformer}.  Existing studies have explored mapping natural language into CAD construction sequences or executable CAD scripts\cite{wu2021deepcad,khan2024text2cad,guan2026cad,li2026towards}. More recent methods such as CAD-Assistant\cite{mallis2025cad}, ToolCAD\cite{gong2026toolcad}, and CADSmith\cite{barkley2026cadsmith} have further investigated tool-augmented CAD agents. These studies indicate that CAD generation has evolved from early sequence generation toward tool-interactive agentic modeling.

However, existing methods still struggle to achieve accurate, reliable, and editable parametric CAD modeling in complex engineering scenarios. The fundamental challenge lies in the mismatch between the probabilistic nature of LLM generation and the deterministic requirements of parametric CAD construction. Because parametric CAD models involve strict geometric, topological, and feature dependencies, any minor error can cause downstream feature failures or invalid topology\cite{camba2016parametric}. Existing methods struggle to bridge this mismatch for three main reasons.

First, most existing methods are strongly dependent on large-scale training or task-specific fine-tuning\cite{wu2021deepcad,khan2024text2cad,guan2026cad}. This paradigm requires large-scale, high-quality paired data between design requirements and CAD construction procedures. Consequently, trained models are often constrained by the training data distribution and are sensitive to variations in natural-language expression and the presence of similar CAD structures in the training data. When user requirements fall outside the training distribution, the generated models may become incomplete, structurally inconsistent, or difficult to execute reliably.

\begin{figure}[htbp]
\centering
\includegraphics[width=\linewidth]{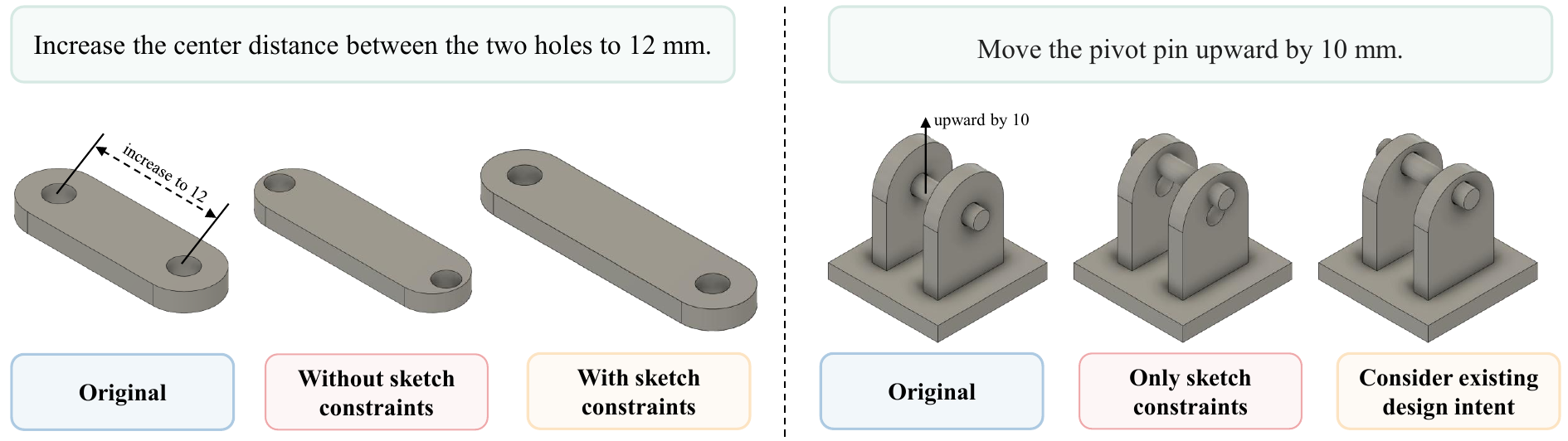}
\caption{Effects of constraints and design-intent modeling on parametric CAD editing. (Left) The imposed tangency and concentricity constraints preserve key geometric relations during CAD editing. (Right) Design-intent modeling propagates the requested modification to dependent features, thereby maintaining the topological consistency of the overall model.}
\label{design_intent}
\end{figure}

Second, existing CAD modeling methods often focus on producing valid geometry, construction sequences, or executable code, while paying insufficient attention to the explicit representation and preservation of design intent in parametric CAD~\cite{wu2021deepcad,khan2024text2cad,guan2026cad,yuan2025cad}. Although these methods may generate visually plausible models, key design relations, topological dependencies and entity-reference relations are often only implicitly embedded in code or operation sequences. In generation, this may lead to requirement drift and reduce consistency with user instruction. This limitation becomes particularly evident in CAD editing, where a valid modification must consider not only the edit instruction but also the design intent already embodied in the original model. Without explicitly modeling such relations, edits may break geometric constraints, invalidate references, or produce structurally inconsistent models~\cite{naya2020role,li2026luban}, as illustrated in \autoref{design_intent}. 

Finally, existing methods do not fully exploit the deterministic capabilities of CAD systems. Modern CAD systems already provide mature modeling and solving capabilities\cite{hoffmann2005constraint,jiushao2025power,autodesk2014fusion}. However, these capabilities are typically exposed through low-level APIs involving fine-grained operations and complex parameters, making them difficult to understand and invoke reliably. Consequently, many existing methods still rely on generative models to maintain geometric consistency rather than leveraging native CAD systems capabilities. This reliance means that even minor errors can cause the entire modeling process to fail.

To address these challenges, we propose ReliCAD, a unified agentic framework that transforms LLM generation into reliable parametric CAD modeling. As illustrated in \autoref{fig:three interaction modes}, ReliCAD supports three interaction modes: Text-to-CAD generation, CAD editing, and iterative generation and refinement. ReliCAD is built around three core mechanisms: explicit design-intent modeling, deterministic CAD-kernel execution, and verification-feedback-driven repair. The main contributions of this paper are as follows:

% ReliCAD organizes the probabilistic LLM generation into an explicit, verifiable and repairable parametric modeling process. As illustrated in \autoref{fig:pipline}, ReliCAD first converts user input into a structured design-intent specification and then identifies the topological dependencies and construction sequence of the target model. It then generates constraint-aware parametric modeling instructions, invokes the Agent-ready API for geometric construction, feature operations, and constraint solving, and records key runtime evidence. Finally, the verification module evaluates the consistency between the resulting model and the user’s design intent based on the design specifications, runtime evidence, and geometric results. When verification fails, ReliCAD analyzes the detected errors, routes them to the corresponding repair operation, and iteratively regenerates and revalidates the model. 

% In this way, ReliCAD mitigates the tension between the probabilistic nature of LLM generation and the deterministic requirements of CAD modeling by reframing CAD modeling as an explicit, verifiable, and repairable parametric modeling workflow. To further validate its effectiveness, we evaluate ReliCAD on parametric CAD generation using the public HistCAD\cite{dong2025histcad} dataset and construct a constraint-aware, multi-granularity CAD editing dataset to assess editing stability, constraint preservation, and design-intent consistency. The main contributions of this paper are as follows:

\begin{enumerate}
\item We propose ReliCAD, a unified agentic framework for parametric CAD modeling. ReliCAD transforms the probabilistic generation of LLMs into an explicit, verifiable, and repairable parametric modeling workflow, improving generalization and modeling stability, and design-intent preservation in complex CAD modeling scenarios.

\item We introduce an explicit design-intent modeling mechanism that incorporates geometric constraints, topological dependencies, and modeling order into the CAD generation and editing process. This mechanism reduces intent drift and improves both design-intent preservation and edit stability of the generated CAD models.

\item We develop an evidence-driven validation and closed-loop repair mechanism that checks consistency between CAD models and design specifications, localizes failures, and performs iterative repair.

\item We construct a multi-granularity CAD editing dataset covering global geometry edits, structural-unit edits, and fine-grained dimensional edits. Together with the HistCAD-based generation benchmark, we systematically evaluate ReliCAD. Experimental demonstrate that ReliCAD outperforms existing methods in terms of generation reliability, design-intent preservation, and edit stability.

\end{enumerate}

\section{Related work}
This section provides a comprehensive review of recent advances in intelligent parametric CAD modeling.

\subsection{Training-based CAD Model Generation}

Training-based CAD model generation methods\cite{wang2025cad,li2025cad,jung2024contrastcad,zhang2025diffusion} usually formulate CAD generation as a process of learning the distribution of existing CAD data. These methods learn the mapping between geometric structures and modeling operations from construction histories or modeling scripts, and then generate new parametric CAD models. Early representative methods, such as DeepCAD\cite{wu2021deepcad} and SkexGen\cite{xu2022skexgen}, represent CAD models as ordered parametric CAD sequences composed of sketching and extrusion operations. They employ transformer-based architectures to learn the distribution of parametric CAD sequences and generate parametric CAD models. Text2CAD\cite{khan2024text2cad} and CAD Translator\cite{li2024cad} further use textual descriptions as conditional inputs and learn the mapping from natural-language design intent to parametric CAD models. Other research directions further investigate generating parametric CAD sequences from B-rep, point clouds, images, and engineering drawings\cite{chen2025img2cad,dupont2024transcad,khan2024cad,ma2024draw,qin2025drawing2cad,rukhovich2025cad,sun2025sketch2seq,you2025img2cad,liang2025cadcl,zhang2026interactive}. Meanwhile, some studies adopt executable modeling code as the target representation for parametric CAD generation\cite{li2026towards,li2026recad}. For example, CADCoder\cite{guan2026cad} and Text-to-CadQuery\cite{xie2025text} employ supervised fine-tuning and reinforcement learning to generate executable CAD scripts.

Training-based CAD model generation methods have substantially advanced parametric CAD generation. However, these methods typically rely on large-scale, high-quality annotated data, which makes their performance sensitive to the training data distribution. When user requirements or geometric structures deviate from the training distribution, these methods often exhibit limited generalization, unstable construction sequences, and even non-executable outputs. Therefore, relying solely on training-based methods is insufficient for stable and generalizable parametric CAD modeling in real industrial applications. In contrast, ReliCAD aims to reduce the reliance on large-scale training data and improve robustness across diverse CAD modeling scenarios.

\subsection{Design Intent Modeling in Parametric CAD}

Design intent refers to the geometric structures, constraint relationships, and feature dependencies involved in the parametric CAD modeling process. It is a key factor that determines model editability and structural consistency\cite{bettig2011geometric}. Understanding and modeling design intent remains an important challenge in parametric CAD model generation\cite{zhou2025status}. Existing studies on explicitly model design intent are still limited. Most of them focus on sketch constraint representation, constraint inference, and constraint-aware dataset construction. For example, SketchGraphs\cite{seff2020sketchgraphs} constructs a large-scale parametric CAD sketch dataset. It represents sketches as graphs composed of geometric primitives and constraint relationships, providing a data foundation for learning geometric relations and constraint patterns in sketches. Vitruvion\cite{seff2021vitruvion} and CadVLM\cite{wu2024cadvlm} further infer geometric constraints from sketches or visual inputs. Casey et al.\cite{casey2025aligning} investigate how automatically generated sketch constraints can be aligned with design intent.  More recently, HistCAD\cite{dong2025histcad} introduces explicit constraints, feature operations, and boundary references into 3D parametric CAD data, further advancing constraint-aware CAD representation and generation.

Although these studies provide important foundations for design intent modeling, they mainly focus on data-level representation. In particular, they emphasize the expression or inference of explicit sketch-level constraints to improve model editability. However, in practical parametric CAD modeling, design intent is not limited to dimensional and geometric constraints in sketches. It also involves topological dependencies among objects and the construction order of features\cite{camba2016parametric}. Different from prior work, ReliCAD explicitly models design intent during the generation and editing process. It incorporates geometric constraints, topological dependencies, and feature construction order into the CAD generation pipeline. This helps reduce intent drift and improves the design intent preservation and editing stability of generated CAD models.

\subsection{LLM-based Agents for CAD Modeling}

Recent advances in large language models have significantly improved their capabilities in code generation, multimodal understanding, and tool use\cite{lu2022learn,lu2022dynamic,chen2021evaluating,schick2023toolformer}. Accordingly, CAD generation is gradually shifting from static sequence generation toward agentic modeling. Early studies began to explore the direct use of general-purpose models such as GPT-4 and GPT-4V to generate CAD scripts or modeling commands\cite{badagabettu2024query2cad,picard2025concept,deng2024investigation}. For example, LLM4CAD \cite{li2025llm4cad} investigates the ability of large language models to generate CAD programming code from multimodal design requirements. Building on this direction, some studies further introduce visual feedback or multimodal information to improve the geometric consistency of generated CAD models\cite{wang2025text}. For instance, CADCodeVerify\cite{alrashedy2025generating} uses a VLM to inspect 3D objects generated from CAD code and iteratively refine the code based on visual feedback. These studies indicate that CAD generation is no longer limited to one-shot static sequence prediction, but has begun to incorporate feedback, refinement, and multi-step reasoning mechanisms.

More recently, researchers have begun to formulate LLMs or vision-language models as CAD agents, enabling them to call external tools, execute CAD operations, and adapt their behavior based on environmental feedback during the modeling process\cite{ataei2026zero}. For example, CAD-Assistant\cite{mallis2025cad} integrates vision-language models with the FreeCAD environment and CAD-specific tools to support CAD question answering, automatic constraint generation, and modeling operations. ToolCAD\cite{gong2026toolcad} further formulates text-to-CAD modeling as a tool-using agent task. These works push CAD generation from simply generating modeling sequences toward executing modeling processes.

However, reliable parametric CAD generation remains an open challenge. Existing CAD agents often use CAD systems mainly as execution environments, while the deterministic capabilities of CAD systems, including constraint solving, parametric updates, and geometric state queries, remain underexplored.  Meanwhile, current feedback and verification mechanisms are typically based on execution success, model renderability, or visual consistency. Although these signals can detect execution failures and apparent shape-level errors, they provide limited support for verifying whether the generated parametric model preserves the geometric constraints, feature relationships, and design intent specified by the design requirements. To overcome these limitations, ReliCAD treats the CAD kernel as a deterministic foundation and introduces evidence-driven validation, thereby transforming the uncertain LLM generation into a constrained, executable, and verifiable parametric CAD modeling process.

\begin{figure*}[htbp]
\centering
\includegraphics[width=\linewidth]{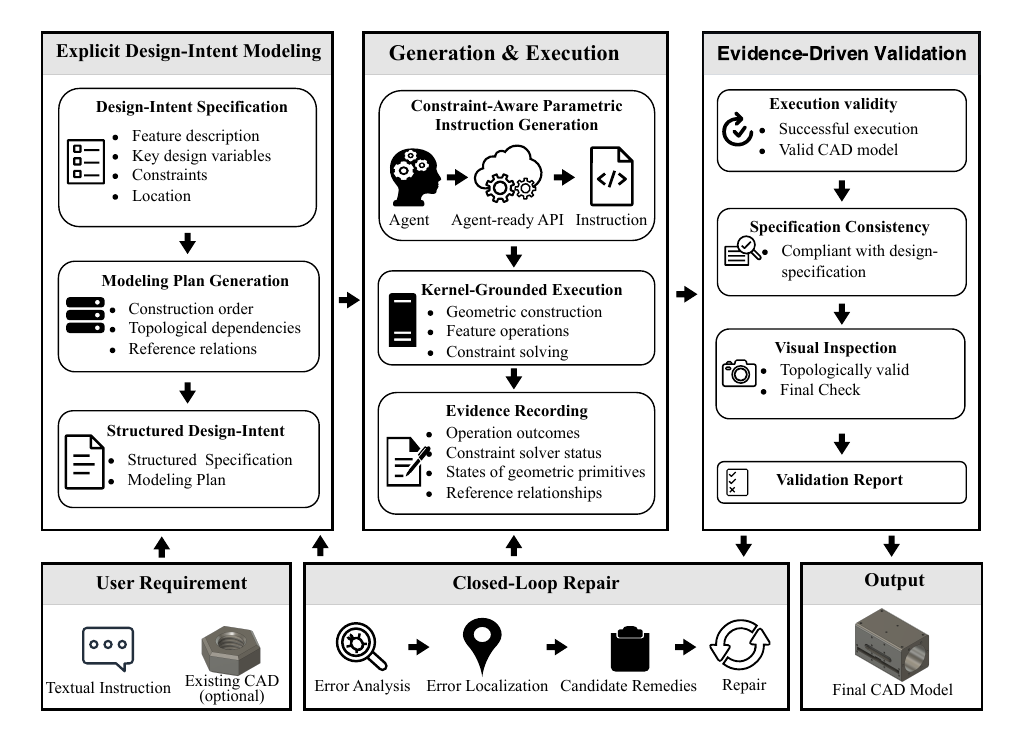}
\caption{Overview of the ReliCAD framework.}
\label{fig:pipline}
\end{figure*}

\section{ReliCAD}
ReliCAD comprises three stages: explicit design-intent modeling, kernel execution via Agent-ready API, and evidence-driven validation with closed-loop repair, transforming LLM generation into reliable parametric modeling.
\subsection{Problem Formulation}
Our goal is to develop a unified agentic framework for parametric CAD model generation and editing, which transforms the black-box probabilistic generation of large language models into a reliable, verifiable, and repairable white-box modeling process. The framework aims to improve the stability of automated CAD modeling, the preservation of design intent, and downstream editability, thus facilitating the deployment of intelligent interactive CAD design in real industrial scenarios.

In this paper, we take natural-language design requirements as input and generate parametric CAD models that satisfy the intended design specifications. Unlike static 3D models that only contain the final geometry, the parametric CAD model $M^{*}$ generated in this work consists of the final geometry $G$, model parameters $\Theta$, constraint relations $C$, and replayable modeling history $H$:
\begin{equation}
M^{*}=(G,\Theta,C,H).
\end{equation}

Based on this definition, given a user design requirement $I$, the parametric CAD model generation task can be formally defined as:
\begin{equation}
M^{*}=\mathcal{F}(I),
\end{equation}
where $\mathcal{F}(\cdot)$ denotes the generation mapping from natural-language design requirements to parametric CAD models. For the editing task, given an existing CAD model $M_0$ and a natural-language editing requirement $I_e$, the system is expected to generate an updated parametric CAD model:
\begin{equation}
M_e^{*}=\mathcal{F}_e(M_0,I_e),
\end{equation}
where $\mathcal{F}_e(\cdot)$ denotes the mapping process for CAD model editing, and $M_e^{*}$ denotes the edited target model.

\subsection{Design Principles and Framework Overview}
ReliCAD’s architecture is driven by the practical demand in industrial scenarios to accurately model user requirements as reliable parametric CAD models. To transform the probabilistic generation of LLMs into a reliable and verifiable parametric modeling process, we establish four design principles. 
% \textbf{Specification-First} and \textbf{Explicit Design-Intent Modeling} ensure that user intent is accurately captured, explicitly represented, and consistently preserved throughout the modeling workflow, while \textbf{Kernel-Grounded Realization} and \textbf{Evidence-Driven Validation and Repair} provide the foundation for the accuracy and reliability of parametric CAD models. The four principles are elaborated below:

\textbf{Specification-First.} Natural-language design requirements are often underspecified and ambiguous. Direct translation of such requirements into CAD modeling code with an LLM may result in missing parameters, misinterpreted structures, or deviations from the intended design objective. To mitigate these issues, user requirements should first be transformed into a structured design specification that explicitly defines the target component, geometric features, dimensional parameters, and design constraints. This specification serves as a shared reference for modeling planning, code generation, and downstream validation, allowing the system to establish a clear and consistent modeling objective prior to execution. 
% By grounding generation in an explicit specification, ReliCAD reduces the uncertainty of  LLM generation and improves the fidelity of the resulting model to the original user intent. 

\textbf{Explicit Design-Intent Modeling.} User design intent should not be left implicit in the textual reasoning or generated code of an LLM. It should be explicitly represented as transferable and inspectable modeling information. Key elements of design intent, including geometric constraints, topological dependencies, and modeling sequences, should be incorporated into the CAD generation process. 
% Thereby, the generated CAD models can better preserve design intent, maintain correct topological relations, and support stable downstream editing.

\textbf{Kernel-Grounded Realization.} Parametric CAD modeling should not rely on an LLM to infer geometric states, constraint relations, or modeling outcomes. Instead, it should be grounded in the native CAD kernel capabilities through the Agent-ready API. 
% Critical operations, including parametric construction, constraint solving, and geometric fact retrieval, should be executed deterministically by the CAD kernel. In this way, probabilistic language reasoning is grounded in deterministic kernel execution, resulting in a more reliable, controllable, and verifiable parametric modeling process.

\textbf{Evidence-Driven Validation and Repair.} The parametric CAD generation process should be grounded in multi-source evidence to verify the consistency between the generated model and the design specification. When validation fails, the system should use this evidence to locate the source of failure and perform a targeted repair. 
% This establishes a closed-loop modeling process that is verifiable, diagnosable, and repairable.

Based on the above design principles, we propose ReliCAD. \autoref{fig:pipline} illustrates the workflow of the framework for modeling parametric CAD models from user requirements. The framework consists of three main stages: 1) \textbf{Explicit Design-Intent Modeling:} ReliCAD converts ambiguous natural-language requirements into explicit design specifications while explicitly modeling the topological dependencies and modeling sequence. 2) \textbf{Kernel-Oriented Parametric Instruction Generation:} ReliCAD generates constraint-aware CAD modeling instructions, invokes the geometric modeling kernel for deterministic execution, and records critical runtime evidence during the modeling process. 3) \textbf{Evidence-Driven Validation and Closed-Loop Repair:} The generated model is validated against the design specification, runtime evidence, and geometric outcomes. When validation fails, ReliCAD performs error analysis and triggers a scheduling mechanism to iteratively repair the model until the design intent is satisfied or the process terminates.

\subsection{Explicit Design-Intent Modeling}
To ensure the consistency between the generated parametric CAD model and the user’s modeling requirements, and to mitigate requirement drift during generation, ReliCAD first performs Explicit Design-Intent Modeling before geometric modeling. This process transforms ambiguous user requirements into structured, explicit information that provides a shared basis for subsequent parametric instruction generation, evidence-driven validation, and closed-loop repair.

\subsubsection{Design-Intent Specification}
% Natural-language design requirements are often characterized by diverse expressions, implicit semantics, and incomplete information. Directly generating CAD modeling code from raw textual descriptions makes the model susceptible to variations in user phrasing and semantic uncertainty. To mitigate this risk, ReliCAD first performs design intent Specification, so that the subsequent modeling process can be grounded in explicit and consistent design evidence.

The core objective of design intent Specification is to map user inputs $I$ into a standardized semantic representation $D$ tailored to parametric CAD modeling.
\begin{equation}
D=\mathcal{S}(I),
\end{equation}
where $D=\{d_1,d_2,\ldots,d_m\}$ denotes the resulting design specification. Each specification item $d_i$ describes a geometric entity or feature together with its parametric attributes and explicit geometric or dimensional constraints. Specifically, ReliCAD maps equivalent modeling requirements expressed in different linguistic forms to a unified design-intent specification, thereby improving robustness to variations in natural-language inputs. Meanwhile, ReliCAD conducts ambiguity resolution to identify issues such as missing critical dimensions, unclear reference objects, and semantic conflicts, and then completes or revises uncertain information accordingly. Through this process, ReliCAD organizes the geometric entities, parametric attributes, spatial relationships, and explicit constraints embedded in ambiguous user requirements into a standardized design specification, which provides a clear target-model description for subsequent modeling stages.

\subsubsection{Modeling Plan Generation}
The structured design specification defines the geometric contents and constraint relations, but does not specify the topological dependencies or their construction order. To prevent modeling failures caused by missing reference objects, incorrect feature sequencing, or broken local structural dependencies, ReliCAD further performs modeling planning on top of the design specification. For generation and editing tasks, the corresponding modeling-planning processes are formulated as:
\begin{equation}
P=\mathcal{P}(D;\mathcal{K}_{\mathrm{CAD}}),
\qquad
P_e=\mathcal{P}_e(M_0,D;\mathcal{K}_{\mathrm{CAD}}),
\end{equation}
where $\mathcal{K}_{\mathrm{CAD}}$ denotes CAD-specific modeling knowledge, $P=[p_1,p_2,\ldots,p_n]$ and $P_e=[p_1^e,p_2^e,\ldots,p_l^e]$ denote ordered sequences of feature construction and modification steps, respectively. Each step specifies the operation type, target feature, required references, and relevant spatial relations.

ReliCAD incorporates CAD modeling priors to identify support dependencies, reference relations, and operation-order dependencies among features. For editing tasks, the planning process additionally analyzes the dependency structure of the existing model $M_0$ to identify which features are affected by the requested modification and which dependent features must be updated to preserve the original design intent. Through this process, ReliCAD converts the requirements into an ordered sequence of construction or modification operations in which all required references are established before use and dependent features are updated coherently during editing, as shown in \autoref{design_intent}.

\subsection{Kernel-Oriented Parametric Instruction Generation}
This section describes how ReliCAD translates the design-intent specification and modeling plan into executable parametric instructions and generates parametric CAD models through a CAD kernel. 
\subsubsection{Agent-ready API}
To provide CAD operations that can be effectively interpreted, invoked, and verified by an agent, ReliCAD introduces an Agent-ready API. Rather than exposing fine-grained native interfaces that are tightly coupled to specific implementations, the API encapsulates CAD-kernel capabilities as structured action units with explicit operation semantics, input parameters, reference objects, preconditions, and returned results. The action space is formally represented as: 
\begin{equation} \mathcal{A} = \mathcal{A}_{\mathrm{sketch}} \cup \mathcal{A}_{\mathrm{constraint}} \cup \mathcal{A}_{\mathrm{feature}} \cup \mathcal{A}_{\mathrm{refinement}} \cup \mathcal{A}_{\mathrm{query}}, \label{eq:agent_ready_action_space} \end{equation} 
where $\mathcal{A}_{\mathrm{sketch}}$, $\mathcal{A}_{\mathrm{constraint}}$, $\mathcal{A}_{\mathrm{feature}}$, $\mathcal{A}_{\mathrm{refinement}}$, and $\mathcal{A}_{\mathrm{query}}$ denote the action subsets for sketch construction, constraint definition, three-dimensional feature generation, local geometric refinement, and geometric query, respectively. By providing unified operation semantics and reference conventions, the Agent-ready API decouples parametric instruction generation from kernel-specific interfaces, allowing the same actions to be mapped to different geometric kernels, including the JiuShao POWER Platform API~\cite{jiushao2025power} and the Autodesk Fusion 360 API~\cite{autodesk2014fusion}. The specific action units currently supported within each subset are summarized in \appref{app:agent_ready_api}.

\subsubsection{Constraint-Aware Parametric Instruction Generation}

\begin{figure}[htbp]
\centering
\includegraphics[width=\linewidth]{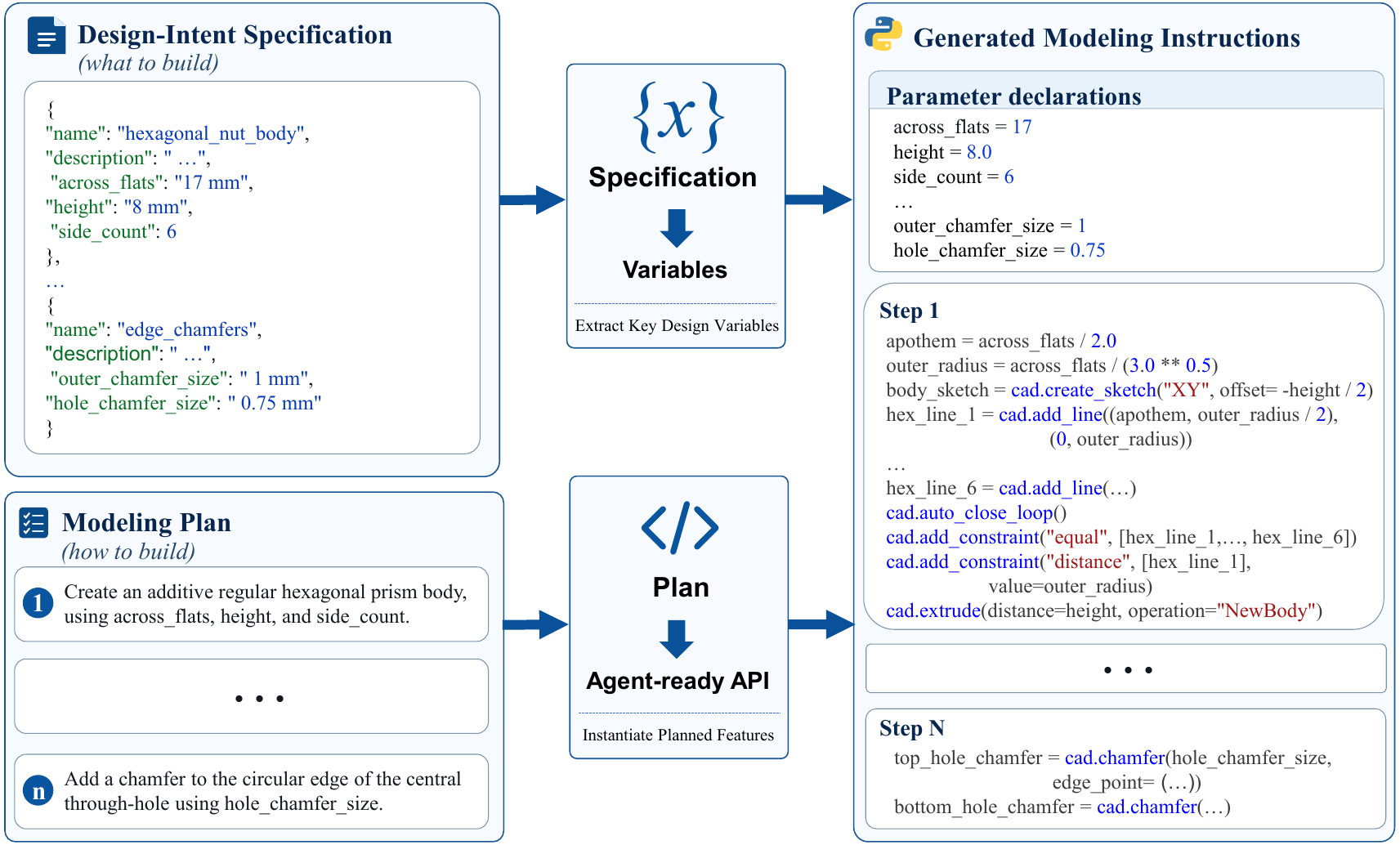}
\caption{Illustration of constraint-aware parametric instruction generation in ReliCAD.}
\label{fig:code_generation}
\end{figure}

Within the unified action space defined by the Agent-ready API, ReliCAD maps the design-intent specification and modeling plan into an ordered sequence of parametric instructions:
\begin{equation}
\Gamma=\mathcal{G}(D,P;\mathcal{A}),
\qquad
\Gamma_e=\mathcal{G}_e(M_0,D,P_e;\mathcal{A}),
\label{eq:parametric_instruction_generation}
\end{equation}
where $\mathcal{G}(\cdot)$ and $\mathcal{G}_e(\cdot)$ denote parametric instruction generation for the generation and editing tasks, respectively.  $\Gamma$ and $\Gamma_e$ denote the resulting ordered instruction sequences. As shown in \autoref{fig:code_generation}, it first extracts the key design parameters defined in the design-intent specification as shared variables for subsequent feature construction. It then instantiates the features defined in the modeling plan by invoking the Agent-ready API for sketch creation, constraint assignment, and three-dimensional feature generation. Operations applied directly to existing bodies, such as filleting, chamfering, and threading, are generated according to the target entities and reference objects specified in the plan. 

During the instruction generation process, ReliCAD adopts a design-intent-oriented constraint strategy. The agent specifies only semantically meaningful constraints, such as parallelism, perpendicularity, concentricity, and equality, while the execution system automatically completes coincidence constraints needed for sketch connectivity and profile closure. This reduces generation the complexity and improves the reliability of parametric modeling.

\subsubsection{Kernel-Grounded Execution and Evidence Recording}
The generated parametric instructions are submitted to the CAD kernel through an adapter layer, which performs geometric construction, Boolean operations, and constraint solving, and returns the corresponding candidate model, runtime evidence, and execution status. The kernel-grounded execution process is formulated as:
\begin{equation}
(\widehat{M},E,s)=\operatorname{Exec}(\Gamma), 
\label{eq:kernel_execution}
\end{equation} 
where $\operatorname{Exec}(\cdot)$ denotes the execution process performed by the CAD kernel, $\widehat{M}$ is the candidate parametric CAD model produced before validation, $E$ denotes the runtime evidence collected during execution, and $s$ represents the execution status. Specifically, $E$ includes operation outcomes, constraint solver status, states of geometric primitives, and reference relationships established or updated during execution. The evidence indicates whether constraints are executed or skipped as redundant, whether constraint solving modifies primitive geometry, and whether referenced entities remain valid after execution. ReliCAD further employs a semantic redundancy-handling mechanism to identify and safely omit semantically valid but redundant constraints in over-constrained models, as detailed in \appref{Semantic Redundancy Handling for Over-Constrained Models}.

\subsection{Evidence-Driven Validation and Closed-Loop Repair}
\label{sec:evidence_validation_repair}
Although the CAD kernel ensures deterministic execution of modeling instructions and constraint solving, the generated results may still deviate from the design requirements due to errors in LLM understanding or generation. Therefore, this section presents ReliCAD’s evidence-driven validation and closed-loop repair mechanism, focusing on the systematic evaluation of generated models and the correction of identified modeling errors.

\subsubsection{Evidence-Driven Validation}
ReliCAD adopts a three-stage evidence-driven validation strategy that integrates complementary evidence from CAD kernel feedback, runtime modeling states, and rendered views. Based on these heterogeneous evidence sources, the system progressively evaluates execution validity, specification consistency, and visual plausibility. The overall validation process is formulated as:
\begin{equation}
\label{eq:validation}
(y,F) = \operatorname{Validate} \left( D,\widehat{M},E,s,\operatorname{Render}(\widehat{M}) \right),  
\end{equation}
where $\operatorname{Render}(\widehat{M})$ denotes the rendered views of the candidate model, $y\in\{0,1\}$ indicates whether all validation stages are passed, and $F$ denotes the corresponding structured diagnostic feedback.

\textbf{Execution validity:} Examines whether the modeling instructions are executed successfully without errors and produce a valid parametric CAD model.

\textbf{Specification consistency:} Assess whether the final model state satisfies each requirement in the design specification. By aligning each specification item with the corresponding runtime evidence from \autoref{eq:kernel_execution}, ReliCAD supports fine-grained, item-wise verification of the generated model and defers inconclusive cases to the subsequent visual validation stage.

\textbf{Visual plausibility:} Uses rendered views to assess the model’s global consistency with the user’s requirements and examine its overall structural integrity and geometric quality. It further revisits specification items that cannot be conclusively verified from runtime evidence.

\subsubsection{Closed-Loop Repair}
When validation detects any failure, specification inconsistency, or visual anomaly, ReliCAD initiates a closed-loop repair process. The system first produces structured validation feedback that localizes the affected modeling step or unmet design requirement, while augmenting the detected issue with the relevant evidence, likely causes, and candidate corrective actions. An error-analysis module then jointly reasons over the validation results, runtime evidence, and CAD domain knowledge to diagnose the underlying cause, determine its impact on related and downstream features, and formulate a targeted repair strategy.

Based on the diagnosed error source, ReliCAD routes the repair request to the corresponding stage of the modeling pipeline. Errors caused by ambiguous or incomplete requirements are redirected to the design-intent specification stage, whereas invalid feature dependencies or construction sequences are handled through modeling replanning. Errors associated with parameter values, constraint definitions, entity references, or individual modeling operations are routed directly to the instruction generation stage for localized correction. This stage routing avoids indiscriminate regeneration of the entire model and confines the modification to the affected design information or modeling steps. At iteration $t$, the targeted repair process is formulated as:
\begin{equation}
\left( D^{(t+1)},P^{(t+1)},\Gamma^{(t+1)} \right) = \operatorname{Repair} \left( D^{(t)},P^{(t)},\Gamma^{(t)},F^{(t)},E^{(t)} \right), 
\end{equation}
where $D^{(t)}$, $P^{(t)}$, and $\Gamma^{(t)}$ denote the design specification, modeling plan, and parametric instruction sequence at iteration $t$. $E^{(t)}$ denotes the runtime evidence and $F^{(t)}$ denotes the corresponding error-diagnosis result, as defined in \autoref{eq:kernel_execution} and \autoref{eq:validation}. According to the diagnosed error source, $\operatorname{Repair}(\cdot)$ updates the affected stage and propagates the revision through its downstream stages, while preserving unaffected design information and modeling steps.

After repair, the revised design specification, modeling plan, or instruction sequence is propagated through the downstream pipeline and re-executed by the CAD kernel, producing updated model states and runtime evidence. The resulting model is then subjected again to the evidence-driven validation procedure. This validation, diagnosis, repair, and re-validation cycle continues until the generated model satisfies the design requirements, thereby transforming uncertain LLM generation into a reliable parametric modeling process. 

\section{Experiments}
On HistCAD, ReliCAD achieves a 99.8\% validity rate and an IoU of 0.8753, outperforming all baselines. Ablation studies confirm the importance of explicit design-intent modeling and validation-driven repair. For CAD editing, ReliCAD also outperforms CAD-Editor on both our multi-granularity editing dataset and a subset sampled from the CAD-Editor dataset.
\subsection{Datasets}
% We evaluate ReliCAD on two tasks: constraint-aware parametric CAD generation and natural-language-driven CAD editing. For generation, existing parametric CAD datasets are used to assess its ability to construct complete parametric models from natural-language specifications. For editing, we construct a multi-granularity CAD editing dataset from existing parametric CAD models. The dataset evaluates ReliCAD’s editing accuracy across varying levels of granularity and its ability to preserve the original design intent.

\subsubsection{Text-to-CAD Test Dataset}
We conduct generation experiments on HistCAD\cite{dong2025histcad}, containing 170,236 parametric CAD models with explicit constraints and four types of textual annotations. From this dataset, we construct a test set of 1,000 samples through stratified weighted sampling, assigning higher sampling probabilities to industrial and geometrically complex models. Further details of the test set, including the sampling, distributional characteristics, and text-annotation selection strategy, are provided in the \appref{app:Generation test dataset construction}.

\subsubsection{Multi-Granularity CAD Editing Dataset}
\label{Multi-Granularity CAD Editing Dataset}
To evaluate text-driven parametric CAD editing, we construct a multi-granularity, constraint-aware editing dataset derived from HistCAD. The dataset covers three editing categories: fine-grained dimensional edits, structural-unit edits, and global geometric edits. Each sample consists of a source parametric CAD model, an editing instruction, and the corresponding target model. The dataset is designed to assess the ability of different methods to accurately perform the specified edits while preserving the original design intent.

The three types of editing pairs are constructed using different transformation strategies. Below, we briefly introduce the construction strategy and corresponding editing scope of each category.

\noindent\textbf{(1) Fine-grained dimensional editing.} This category targets precise modifications to local geometric parameters. We first randomly modify a dimensional or positional parameter of a sketch primitive in the modeling sequence. A geometric constraint solver is then used to recompute the positions of related primitives according to the existing constraints. Finally, the modified modeling sequence is subjected to geometric-validity and executability checks, and samples that pass these checks are retained as fine-grained dimensional editing pairs.

\noindent\textbf{(2) Structural-unit editing.} This category focuses on modifications to structural units, including loops, faces, sketches, and extrusion features. Specifically, a target structural unit in the CAD modeling sequence is randomly masked, and a fine-tuned LLM is used to reconstruct the masked content conditioned on the remaining sequence context, thereby producing editing pairs with local structural differences\cite{zhang2025flexcad}.

\noindent\textbf{(3) Global geometric editing.} This category targets the entire CAD model or a large-scale local region and covers geometric editing operations such as duplication, translation, deletion, and rotation. These operations are used to construct editing tasks involving large-scope geometric transformations and structural reorganization.

Across all three editing categories, the original modeling sequence is treated as the target output, while the transformed sequence serves as the source model to be edited. This reverse construction strategy preserves the distributional characteristics and modeling conventions of real parametric CAD sequences in the target output, avoiding artifacts introduced by automatically synthesized targets. For each resulting source–target sequence pair, an editing instruction is generated by GPT-5.5 based on the differences between the two CAD models.

Following data construction and annotation, we sample 100 instances from each of the fine-grained dimensional, structural-unit, and global geometric editing categories, resulting in a multi-granularity editing test set of 300 samples. 

\subsubsection{CAD-Editor Test Set}
CAD-Editor represents CAD parameters through discrete quantization, which makes it impossible to precisely represent specific dimensional values. In addition, its training instructions do not include dimensional information, resulting in a distribution mismatch with our editing test set. To ensure a fair evaluation and avoid underestimating CAD-Editor based solely on our test set, we further sample 100 instances from its official test set and conduct separate comparative experiments on the two test sets.

\subsection{Experimental Setup}
\subsubsection{Baseline}
\textbf{Text-to-CAD Baselines.} We compare ReliCAD against four dedicated Text-to-CAD methods and one general-purpose code model. Based on their output representations, Text-to-CadQuery\cite{xie2025text} and CAD-Coder\cite{guan2026cad} generate executable CadQuery code, whereas CADFusion\cite{wang2025text} and HistCAD\cite{dong2025histcad} produce parametric CAD modeling sequences. For the general-purpose baseline, we use GPT-5.5 in Codex with the reasoning effort set to xhigh and adopt four-shot in-context prompting. Codex generates modeling code using the same Agent-ready API as ReliCAD and is allowed to autonomously select and invoke available skills during inference. The detailed inputs and prompts are provided in \appref{app:codex_baseline_prompt}. 
\textbf{CAD Editing Baseline.} We compare ReliCAD with CAD-Editor\cite{yuan2025cad}, which takes a quantized parametric modeling sequence and a natural-language editing instruction as input and outputs the edited modeling sequence. As described in Section~\ref{Multi-Granularity CAD Editing Dataset}, the comparison is conducted on two test sets: our constructed editing dataset and a subset sampled from the CAD-Editor test set.

\subsubsection{Evaluation Metrics}
We evaluate parametric CAD generation and editing using Validity Rate (VR), Intersection over Union (IoU), Chamfer Distance (CD), Jensen--Shannon Divergence (JSD), and Repair Iterations. VR measures the proportion of outputs that can be successfully executed by the CAD kernel to produce valid geometric models. IoU evaluates volumetric overlap, while CD and JSD measure surface-level geometric discrepancy and spatial-distribution discrepancy, respectively. Repair Iterations records the number of closed-loop repair attempts required to obtain the final output.

IoU, CD, and JSD are computed only over valid generations, with failed samples excluded from the aggregate statistics. Consequently, these metrics should be interpreted jointly with VR, as methods with low VR may exhibit overestimated geometric quality on the remaining valid samples. Detailed definitions and computation procedures are provided in \appref{app:evaluation_metrics}.

\subsubsection{Implementation Details}
We use LangGraph to orchestrate the CAD modeling workflow of ReliCAD, with GPT-5.5 serving as the base model. To improve output stability, the temperature is set to 0 for all inference processes. ReliCAD adopts the Agent-ready API defined in this paper as the unified modeling interface. This interface supports backend calls to both Autodesk Fusion 360 and the JiuShao CAD kernel. In our experiments, to ensure a consistent execution environment and evaluation pipeline, all experiments use the Fusion 360 API as the CAD execution backend. When a generated result fails validation, ReliCAD performs closed-loop repair based on the validation feedback and re-executes the repaired modeling sequence. To avoid unbounded reasoning, the maximum number of repair iterations is set to three in all experiments.

\subsection{Main Results}
\subsubsection{CAD Generation Results}
\label{sec:CAD Generation Results}

\begin{table}[htbp]
\caption{Quantitative results on generation task}
\label{tabe:generate_compaire}
\centering
\begin{tabular}{lcccc} 
\hline
Method            & VR↑               & IoU↑             & Med. CD↓        & JSD↓              \\ 
\hline
Text2CadQuery     & 78.2\%            & 13.80            & 1.83            & 84.74             \\
CADFusion         & 76.9\%            & 19.23            & 1.09            & 79.55             \\
CAD-Coder         & 91.6\%            & 23.18            & 1.24            & 75.74             \\
HistCAD           & 97.6\%            & 48.11            & 0.36            & 48.81             \\
Codex-5.5-Xhigh   & 98.5\%            & 85.76            & 0.01           & 16.22             \\
\textbf{ReliCAD } & \textbf{ 99.8\% } & \textbf{ 87.53$^{\dagger}$ } & \textbf{ 0.01 } & \textbf{ 15.13 }  \\
\hline
\end{tabular}

{\footnotesize\justifying
VR is reported as a percentage. IoU, CD,
and JSD are multiplied by 100 and computed only over valid
generations. $^{\dagger}$ indicates a statistically significant improvement over the best-performing baseline at $p<0.001$, as detailed in \autoref{sec:CAD Generation Results}.\par}
\end{table}

\begin{figure*}[htbp]
\centering
\includegraphics[width=\linewidth]{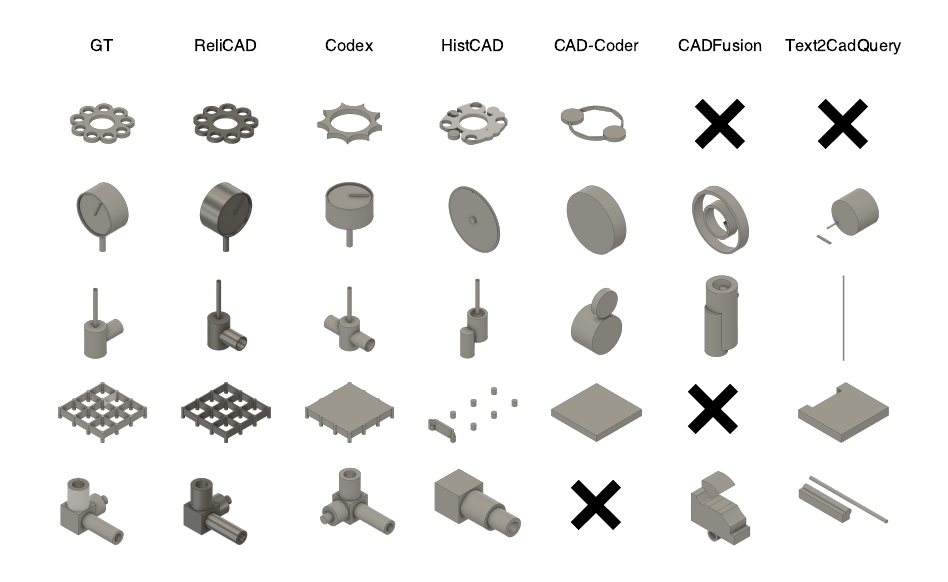}
\caption{Qualitative results on parametric CAD model generation task.}
\label{fig:Quantative results on generation task}
\end{figure*}

\autoref{tabe:generate_compaire} summarizes the quantitative results on the generation task. ReliCAD significantly outperforms the specialized training-based methods across all evaluation metrics. Due to limitations in training-data distribution and model representation capacity, these methods perform well on simple samples but struggle to accurately generate the complex geometric structures present in the test set. Codex-5.5-Xhigh serves as a strong baseline by using the Agent-ready API developed in this work, together with Skill invocation, four in-context examples, and external CAD modeling knowledge, thereby providing substantial modeling-reasoning and tool-use capabilities. Nevertheless, ReliCAD achieves consistently better performance in terms of VR, IoU, and JSD. To further assess whether the improvement over Codex-5.5-Xhigh is statistically significant, we assign an IoU score of zero to failed generations and conduct a two-sided paired $t$-test over all 1,000 paired samples. The results show that ReliCAD achieves a statistically significant improvement in IoU over Codex-5.5-Xhigh ($t(999)=4.994, p=6.98\times10^{-7}$), indicating that the observed improvement is unlikely to be explained by sampling variation. Overall, these results demonstrate the effectiveness of ReliCAD in improving the reliability and geometric consistency of CAD generation, and further suggest that explicit design-intent modeling, evidence-driven validation, and closed-loop repair jointly contribute to more reliable parametric CAD generation.

\autoref{fig:Quantative results on generation task} presents the generation results of different methods on the HistCAD dataset. The CAD models generated by ReliCAD exhibit high consistency with the ground-truth models in terms of global structure, local features, and their topological relationships. In comparison, although Codex generates models that broadly resemble the targets, noticeable deviations remain in feature geometry, dependency relationships, and relative positioning. Other specialized methods are more prone to excessive structural simplification, missing critical features, incorrect component relationships, or invalid geometry. These results demonstrate that ReliCAD can more accurately interpret complex modeling requirements and generate structurally complete parametric CAD models.

\subsubsection{CAD Editing Results}

\begin{table}[htbp]
\caption{Quantitative results on editing task}
\label{tabe:edit_compaire}
\centering
\begin{tabular}{lccccc} 
\hline
Method      & VR↑    & IoU↑  & Avg. CD↓ & JSD↓   \\ 
\hline
CAD-Editor  & 73.7\% & 49.01 & 0.86     & 42.45  \\
ReliCAD     & 99.7\% & 90.61 & 0.01     & 13.01  \\
\hdashline[1pt/1pt]
CAD-Editor* & 94\%   & 58.75 & 0.32     & 41.71  \\
ReliCAD*    & 98\%   & 63.38 & 0.29     & 31.79  \\
\hline
\end{tabular}

{\footnotesize\raggedright
The first two rows are obtained on our multi-granularity editing dataset, whereas $^{*}$ denotes evaluation on a subset sampled from the CAD-Editor test set. VR is reported as a percentage. IoU, CD, and JSD are multiplied by 100 and computed only over valid editing outputs.\par}
\end{table}

\begin{figure*}[htbp]
\centering
\includegraphics[width=\linewidth]{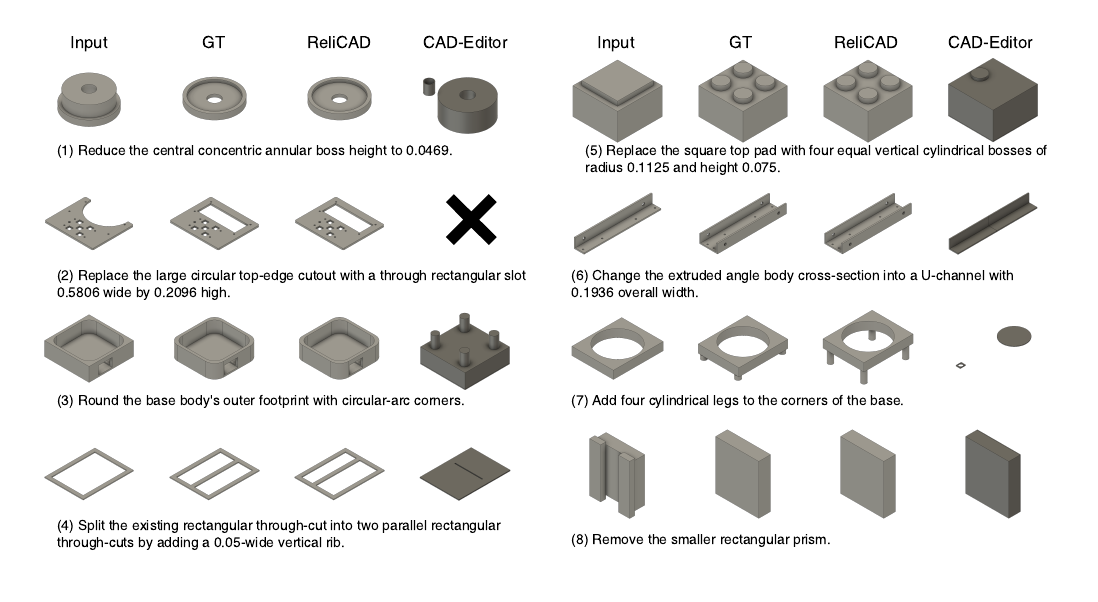}
\caption{Qualitative results on the parametric CAD model editing task.}
\label{fig:Qualitative results on parametric CAD model editing task.}
\end{figure*}

We evaluate the methods on both the multi-granularity editing dataset constructed in this work and a subset sampled from the CAD-Editor test set. The results are reported in Table~\ref{tabe:edit_compaire}. ReliCAD outperforms CAD-Editor across all metrics on both datasets, demonstrating that explicit design-intent modeling, validation-driven repair, and deterministic kernel execution effectively improve the reliability and geometric consistency of CAD editing.

CAD-Editor performs better on its own test subset than on our dataset. This improvement can be attributed primarily to the relative simplicity of the sampled models and the close alignment of the corresponding editing tasks with its training distribution. By contrast, our dataset contains more complex edits with explicit dimensional instructions. Since CAD-Editor is trained without exact dimensional information, it struggles to faithfully execute such instructions. ReliCAD achieves lower scores on the CAD-Editor subset than on our dataset largely because the instructions in that subset are often underspecified and may admit multiple valid editing outcomes, whereas evaluation is conducted against a single reference model. Nevertheless, ReliCAD consistently outperforms CAD-Editor on both datasets, demonstrating greater reliability and generalization across diverse CAD editing scenarios.

\autoref{fig:Qualitative results on parametric CAD model editing task.} presents representative results across different types of CAD editing tasks. ReliCAD accurately performs dimensional adjustments, feature replacement, and feature addition or removal, while preserving the geometry and feature dependencies of unaffected regions. By contrast, although CAD-Editor can handle simple edits, it often produces incomplete modifications, loss of existing features, disconnected components, or even invalid models when precise dimensions or complex structural relationships are involved. These results demonstrate that ReliCAD is more reliable in following editing instructions, preserving the original design intent, and propagating structural changes.

\subsection{Ablation Studies}

\begin{table*}[htbp]
\caption{Quantitative ablation results}
\label{table:ablation results}
\centering
\begin{tabular}{lccccc} 
\hline
Method                     & VR↑    & IoU↑  & Avg. CD↓ & JSD↓  & Avg. Repair Iterations ↓  \\ 
\hline
w/o S\&P        & 96.1\% & 84.33 & 163.90   & 18.11 & 0.687                     \\
Basic V\&R & 92.3\% & 85.91 & 30.77    & 16.42 & 0.485                     \\
Full ReliCAD               & 99.8\% & 87.53 & 0.12     & 15.13 & 0.524                     \\
\hline
\end{tabular}\\
{\footnotesize
VR is reported as a percentage. IoU, CD,
and JSD are multiplied by 100 and computed only over valid
generations.\par}
\end{table*}

\begin{figure}[htbp]
\centering
\includegraphics[width=\linewidth]{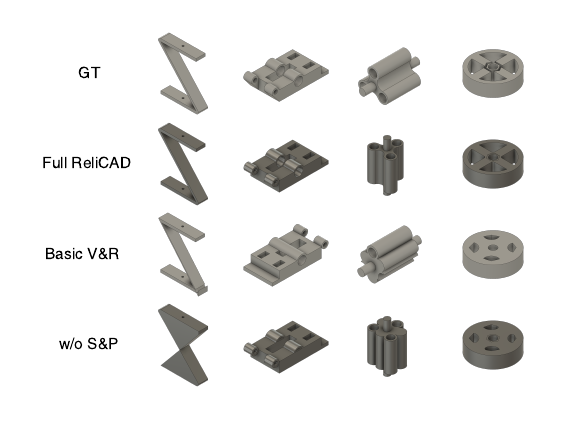}
\caption{Qualitative ablation results.}
\label{fig:Qualitative ablation results.}
\end{figure}

To isolate the contributions of explicit design-intent modeling and evidence-driven validation and repair, we construct two ablated variants of the full ReliCAD framework. \textit{w/o S\&P} removes design-intent specification and modeling planning, preventing the system from explicitly representing geometric relations, feature dependencies, and construction order. \textit{Basic V\&R} replaces the evidence-driven, multi-level validation mechanism with visual validation alone. Upon detecting an anomaly, this variant first repairs the parametric instructions and falls back to intent modeling if the repair fails.

\autoref{table:ablation results} reports the quantitative results. The full ReliCAD achieves the best performance across all metrics, confirming the effectiveness of both core components. Removing design-intent specification and modeling planning degrades model validity and geometric consistency, indicating that the lack of explicit design-intent constraints increases errors in feature relations and topological structure and allows them to propagate into subsequent repair stages. The pronounced drop in VR under Basic V\&R primarily reflects its weaker repair capability. Without runtime-evidence-based diagnosis and error routing, the agent cannot accurately localize failure sources or generate targeted corrections, leaving many modeling errors unresolved within the repair budget. Although this variant requires slightly fewer repair attempts than the full ReliCAD, this does not indicate greater repair efficiency. Instead, its lower repair count mainly results from errors that remain undetected and consequently do not trigger the repair process.

\autoref{fig:Qualitative ablation results.} provides qualitative comparisons. Without design-intent specification and modeling planning, the generated models often exhibit incorrect feature connectivity and global structure. Although Basic V\&R can produce results that appear visually plausible, they may still retain missing features, incorrect connections, or dimensional inconsistencies. In contrast, the full ReliCAD better preserves the target geometry and topology. These results demonstrate that explicit design-intent modeling prevents structural errors during generation, while evidence-driven validation and repair identifies and corrects residual errors, jointly improving the reliability of parametric CAD generation.

\section{Conclusion and Future Work}
We presented ReliCAD, a unified agentic framework that transforms uncertain LLM generation into reliable parametric CAD modeling through explicit design-intent modeling, Agent-ready API, and evidence-driven validation with closed-loop repair. On the HistCAD generation benchmark, ReliCAD significantly outperforms existing training-based baselines. On our multi-granularity editing dataset, ReliCAD effectively preserves design intent during editing. Ablation studies further confirm the contributions of explicit design-intent modeling and evidence-driven validation with closed-loop repair.

Future work will extend ReliCAD to multimodal inputs, including text, sketches, and reference images. We will also extend the framework to complex assembly modeling and integrate domain-specific knowledge, such as manufacturing constraints, functional requirements, and engineering specifications, to improve the engineering feasibility and practical applicability of the generated models. Furthermore, we will explore online reinforcement learning to further enhance ReliCAD’s modeling capabilities and reduce repair iterations.

\bibliographystyle{IEEEtran}
\bibliography{sample}

\appendices
% \section{Agent-ready API Action Space}
% \label{app:agent_ready_api}

\section{Agent-ready API Action Space}
\label{app:agent_ready_api}

\begin{figure*}[t]
    \centering
    \includegraphics[width=\textwidth]{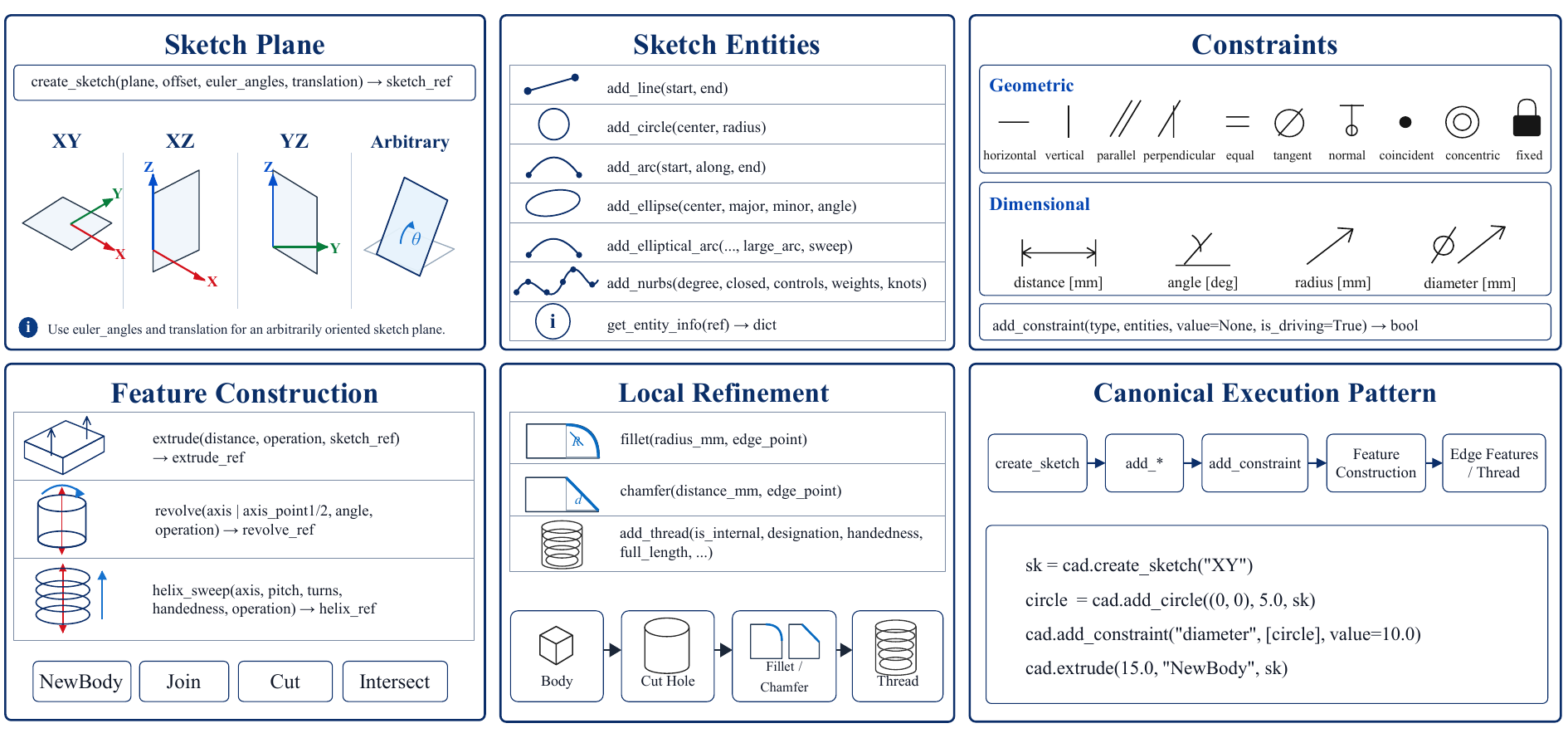}
    \caption{Overview of the Agent-ready API.}
    \label{fig:agent_ready_api_overview}
\end{figure*}

This appendix summarizes the action units currently supported by the Agent-ready API. Following the action-space decomposition defined in \autoref{eq:agent_ready_action_space}, the supported actions are organized into five subsets: sketch construction, constraint, feature construction, local geometric refinement, and geometric query. \autoref{fig:agent_ready_api_overview} provides a visual overview of the representative action units and illustrates how they are composed in a canonical modeling workflow, while \autoref{tab:agent_ready_api_actions} gives the complete list of actions included in each subset. Auxiliary operations for model export, validation, and session management are listed separately, as they are not part of the formal modeling action space.

\begin{table*}[htbp]
\caption{Action units currently supported by the Agent-ready API}
\label{tab:agent_ready_api_actions}
\centering
\begin{tabular}{ll} 
\hline
Action Type          & Supported Actions                                                                                 \\ 
\hline
Sketch Construction  & create\_sketch, add\_line, add\_circle, add\_arc, add\_ellipse, add\_elliptical\_arc, add\_nurbs  \\
Constraint           & auto\_close\_loop, add\_constraint                                                                 \\
Feature Construction & extrude, revolve, helix\_sweep                                                                    \\
Local Refinement     & fillet, chamfer, add\_thread                                                                      \\
Geometric Query      & get\_entity\_info                                                                                 \\
Auxiliary Operations & export\_step, export\_view, validate\_step\_file,  ping, clear, close                              \\
\hline
\end{tabular}
\end{table*}

Complex modeling procedures are realized by composing these modeling action units according to the generated modeling plan. Kernel-specific adapters map each action unit to the corresponding implementation provided by the underlying
CAD kernel.

\section{Semantic Redundancy Handling for Over-Constrained Models}
\label{Semantic Redundancy Handling for Over-Constrained Models}

LLM-generated instruction sequences may contain constraints that are semantically valid but redundant with the current geometric configuration. To prevent such constraints from unnecessarily interrupting kernel execution, ReliCAD employs a semantic redundancy-handling mechanism for over-constrained models. When the constraint solver reports an over-constraint, the execution system examines whether the geometric relation encoded by the newly introduced constraint is already satisfied by the target entities. If the relation already holds, the constraint is classified as redundant and omitted, allowing execution to continue. An error is raised only when the new constraint is inconsistent with the current geometric configuration or conflicts with existing constraints.

\section{Generation test dataset construction}
\label{app:Generation test dataset construction}

Our generation experiments are conducted on HistCAD\cite{dong2025histcad}, a large-scale dataset that integrates models from DeepCAD\cite{wu2021deepcad}, SketchGraphs\cite{seff2020sketchgraphs}, and the Fusion 360 Gallery\cite{willis2021fusion360}, together with professionally authored CAD models from real-world industrial applications. HistCAD contains 170,236 parametric CAD models and covers not only standard sketching and extrusion operations, but also more advanced features, including revolutions, helical sweeps, fillets, and chamfers. It also explicitly represents geometric and dimensional constraints within sketches. Each model is accompanied by four types of textual annotations at different levels of granularity: a detailed Natural Language Transcription (NLT), a modeling-process description, a geometric-structure description, and a functional-type description.

\begin{figure*}[htbp]
    \centering

    \subfloat[Distribution of dataset sources.%
    \label{fig:dataset_source}]{
        \includegraphics[width=0.39\textwidth]
        {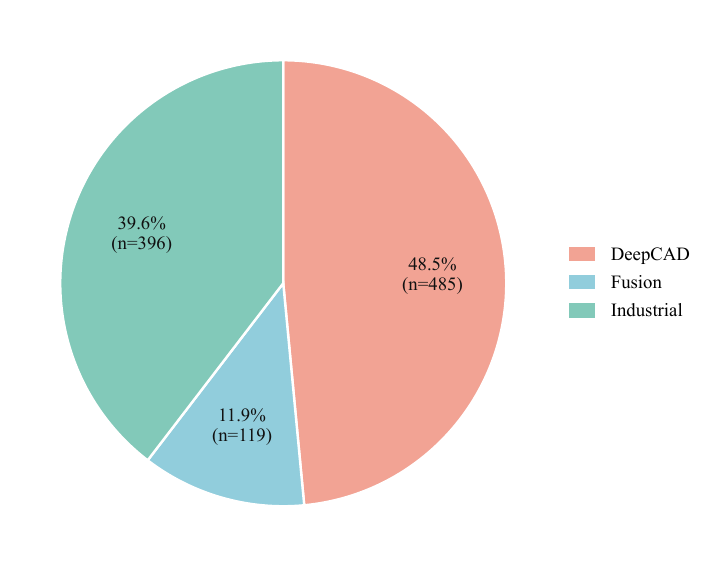}
    }%
    \hfill
    \subfloat[Distribution of modeling operations per sample.%
    \label{fig:operation_distribution}]{
        \includegraphics[width=0.58\textwidth]
        {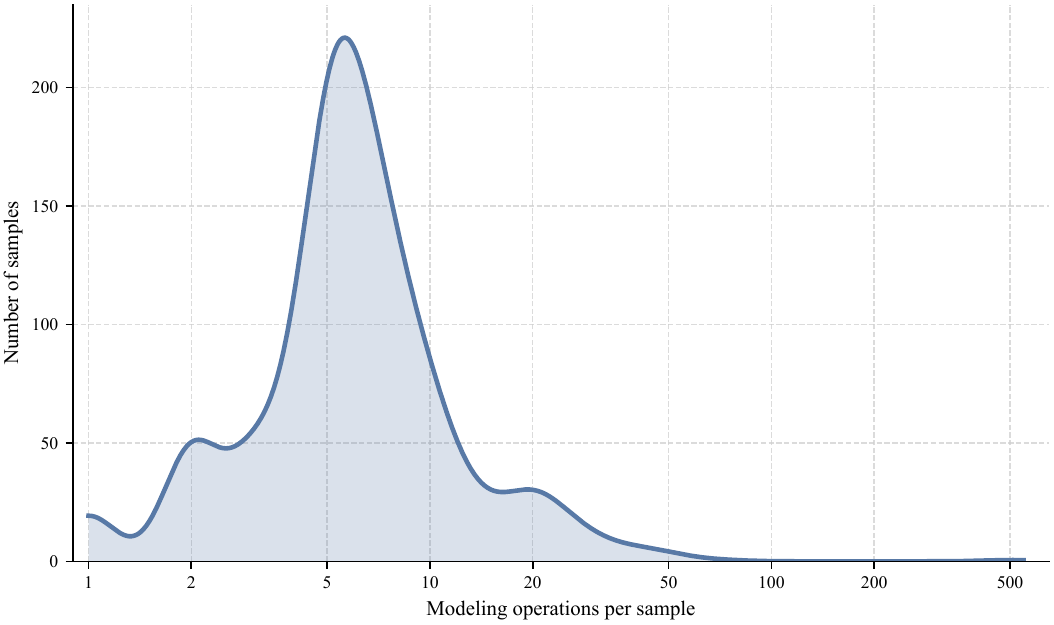}
    }

    \caption{Statistics of the evaluation dataset.}
    \label{fig:dataset_statistics}

\end{figure*}

Based on this dataset, we construct a generation test set of 1,000 samples using stratified weighted sampling. To better reflect practical engineering scenarios, we increase the sampling weight of the industrial subset, thereby raising the proportion of professionally authored CAD models. Within the academic subset, models from the Fusion 360 Gallery are assigned a higher weight to improve source diversity. We further favor high-complexity samples, allowing the test set to cover more models involving multi-step construction, intricate local features, and inter-feature dependencies. Therefore, the resulting test set better represents the complexity of real-world parametric CAD modeling, with its sample distribution shown in \autoref{fig:dataset_statistics}.

Among the four annotation types provided by HistCAD, we use the geometric-structure descriptions as natural-language modeling requirements. Functional-type descriptions generally specify only the object category or intended engineering function, without providing sufficient geometric descriptions to uniquely define the target shape. As a result, the same description may correspond to multiple plausible geometries. In contrast, NLTs and modeling-process descriptions expose detailed operation sequences and feature locations that differ substantially from the design requirements typically provided by users. Using such inputs would reduce the task to translating natural-language operation sequences into parametric code or modeling instructions, rather than requiring a method to independently plan the construction process and generate the CAD model. Geometric-structure descriptions offer a more suitable balance between information completeness and input naturalness: they specify the principal geometry of the target model without revealing its construction procedure, making them well suited for evaluating natural-language-driven parametric CAD generation.

\section{Prompt for the Codex Baseline}
\label{app:codex_baseline_prompt}

\begin{figure*}[htbp]
    \centering
    \includegraphics[width=\textwidth]{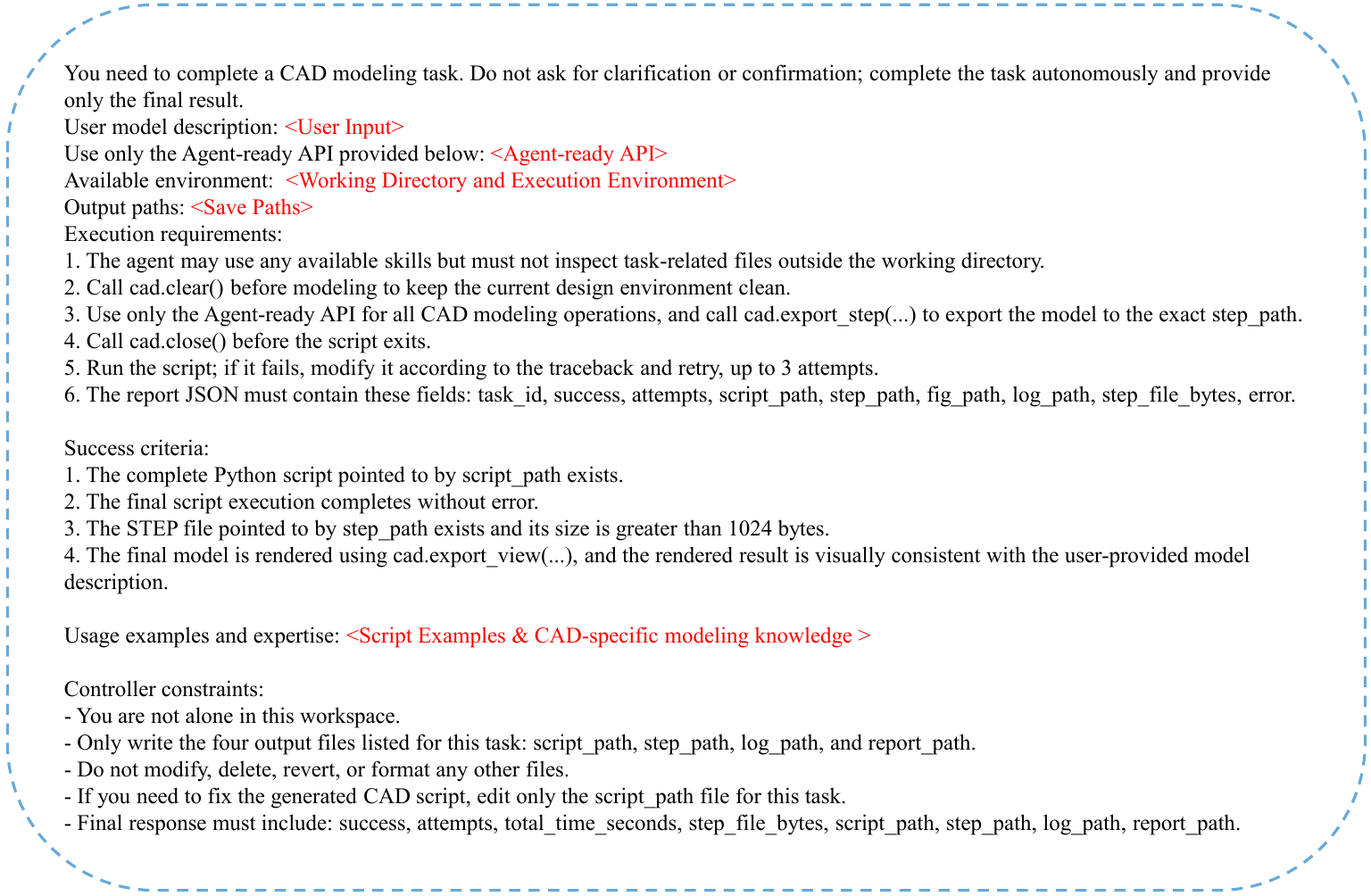}
    \caption{Prompt used for the Codex-5.5-Xhigh baseline.}
    \label{fig:codex_baseline_prompt}
\end{figure*}

The complete prompt used for the Codex-5.5-Xhigh baseline is shown in \autoref{fig:codex_baseline_prompt}. Codex uses GPT-5.5 with the reasoning effort set to xhigh and four fixed in-context examples. For each test sample, it receives the original user description and the same Agent-ready API used by ReliCAD. Codex is required to generate and execute the modeling script, export the
resulting STEP model, and revise the script based on errors, with at most three attempts. All prompt and execution settings are kept consistent across test samples.

\section{Evaluation Metrics}
\label{app:evaluation_metrics}

This appendix provides the detailed definitions and computation
procedures of the evaluation metrics used for parametric CAD generation
and editing.
 
\textbf{Validity Rate (VR).}
VR measures the proportion of test cases for which the final output constitutes a valid CAD model. VR is defined as
\begin{equation}
\operatorname{VR}
=
\frac{N_{\mathrm{valid}}}{N},
\end{equation}
where $N_{\mathrm{valid}}$ is the number of test cases whose final outputs can be successfully executed by the CAD kernel to produce valid geometric models, and $N$ is the total number of test cases. A higher VR indicates that the method is more reliable in producing valid CAD models.

\textbf{Intersection over Union (IoU).}
IoU measures the volumetric similarity between a generated CAD model and its reference model. Let $\mathcal{G}$ and $\mathcal{R}$ denote the generated and reference models, respectively. IoU is defined as:
\begin{equation}
\operatorname{IoU}(\mathcal{G},\mathcal{R})
=
\frac{
\operatorname{Vol}(\mathcal{G}\cap\mathcal{R})
}{
\operatorname{Vol}(\mathcal{G}\cup\mathcal{R})
},
\end{equation}
where $\operatorname{Vol}(\mathcal{G}\cap\mathcal{R})$ is the volume of their intersection, and $\operatorname{Vol}(\mathcal{G}\cup\mathcal{R})$ is the volume of their union. IoU ranges from 0 to 1, where a higher value indicates greater volumetric agreement. An IoU of 1 corresponds to perfect overlap, whereas an IoU of 0 indicates no overlap.

\textbf{Chamfer Distance (CD).}
CD measures the surface-level geometric discrepancy between a generated CAD model and its reference model. For each model, we uniformly sample 10,000 points from its surface. Let $\mathcal{P}$ and $\mathcal{Q}$ denote the point sets sampled from the generated model $\mathcal{G}$ and the reference model $\mathcal{R}$, respectively. CD is defined as:
\begin{equation}
\operatorname{CD}(\mathcal{P},\mathcal{Q})
=
\frac{1}{|\mathcal{P}|}
\sum_{\mathbf{p}\in\mathcal{P}}
\min_{\mathbf{q}\in\mathcal{Q}}
\frac{\|\mathbf{p}-\mathbf{q}\|_2^2}{s_{\mathcal{R}}^2}
+
\frac{1}{|\mathcal{Q}|}
\sum_{\mathbf{q}\in\mathcal{Q}}
\min_{\mathbf{p}\in\mathcal{P}}
\frac{\|\mathbf{q}-\mathbf{p}\|_2^2}{s_{\mathcal{R}}^2},
\end{equation}
where $s_{\mathcal{R}}$ denotes the scale of the reference model, defined as the diagonal length of its axis-aligned bounding box. We normalize CD by the reference-model scale to reduce bias from size variations across test samples. Since the generated and reference models are not normalized independently, scale errors in the generated model remain penalized and may lead to extremely large CD values. A lower CD indicates closer surface-level agreement between the generated and reference models.

\textbf{Jensen--Shannon Divergence (JSD).}
JSD measures the discrepancy between the spatial distributions of a generated CAD model and its reference model. Following the point-based evaluation protocol, we first sample surface points from both models and voxelize them within a shared bounding volume. The voxelized point distributions are then normalized into discrete probability distributions. Let $\mathbf{p}_{\mathcal{G}}$ and $\mathbf{p}_{\mathcal{R}}$ denote the spatial occupancy distributions of the generated model $\mathcal{G}$ and the reference model $\mathcal{R}$, respectively. JSD is defined as:
\begin{equation}
\operatorname{JSD}(\mathbf{p}_{\mathcal{G}}, \mathbf{p}_{\mathcal{R}})
=
\frac{1}{2}
D_{\mathrm{KL}}
\left(
\mathbf{p}_{\mathcal{G}}
\middle\|
\mathbf{m}
\right)
+
\frac{1}{2}
D_{\mathrm{KL}}
\left(
\mathbf{p}_{\mathcal{R}}
\middle\|
\mathbf{m}
\right),
\end{equation}
\begin{equation}
\mathbf{m}
=
\frac{1}{2}
\left(
\mathbf{p}_{\mathcal{G}}
+
\mathbf{p}_{\mathcal{R}}
\right),
\end{equation}
where $D_{\mathrm{KL}}(\cdot\|\cdot)$ denotes the Kullback--Leibler divergence. A lower JSD indicates that the generated model has a spatial distribution more consistent with the reference model.

\textbf{Repair Iterations.}
Repair Iterations measures the number of closed-loop repair attempts required to obtain the final output. An iteration is counted when the system diagnoses a validation failure. If the initial output passes validation, the number of repair iterations is zero. In our experiments, the maximum number of repair iterations is set to three to prevent unbounded reasoning. A lower value indicates a more efficient generation or editing process.

For IoU, CD, and JSD, each generated model and its reference model are rotation-aligned and center-aligned before evaluation. These geometry-similarity metrics are computed only on valid outputs, while failed samples are excluded from the mean and median calculations. Consequently, IoU, CD, and JSD should be interpreted jointly with VR, as methods with low VR may exhibit overestimated geometric quality on the remaining valid samples.

\end{document}